# Differentiated Thyroid Cancer Recurrence Classification Using Machine Learning Models and Bayesian Neural Networks with Varying Priors: A SHAP-Based Interpretation of the Best Performing Model

HMNS Kumari[1#], HMLS Kumari[2], and UMMPK Nawarathne[3]
[1]Faculty of Information Technology and Communication Sciences, Tampere University, Finland
[2]Computing Centre, Faculty of Engineering, University of Peradeniya, Sri Lanka
[3]Faculty of Computing, Sri Lanka Institute of Information Technology, Sri Lanka

[#]nadeeshashyamikumari@gmail.com

**ABSTRACT** Differentiated thyroid cancer (DTC) recurrence is a major public health concern, requiring classification and predictive models that are not only accurate but also interpretable and uncertainty-aware. This study introduces a comprehensive framework for DTC recurrence classification using a dataset containing 383 patients and 16 clinical and pathological variables. Initially, 11 machine learning (ML) models were employed using the complete dataset, where the Support Vector Machines (SVM) model achieved the highest accuracy of 0.9481. To reduce complexity and redundancy, feature selection was carried out using the Boruta algorithm, and the same ML models were applied to the reduced dataset, where it was observed that the Logistic Regression (LR) model obtained the maximum accuracy of 0.9611. However, these ML models often lack uncertainty quantification, which is critical in clinical decision making. Therefore, to address this limitation, the Bayesian Neural Networks (BNN) with six varying prior distributions, including Normal (0,1), Normal (0,10), Laplace (0,1), Cauchy (0,1), Cauchy (0,2.5), and Horseshoe (1), were implemented on both the complete and reduced datasets. The BNN model with Normal (0,10) prior distribution exhibited maximum accuracies of 0.9740 and 0.9870 before and after feature selection, respectively. As the BNN model with (0,10) after feature selection outperformed other models, it was chosen as the best-performing model for DTC recurrence classification. This model was further analyzed using epistemic and aleatoric uncertainty, reflecting the model's confidence in its prediction. In addition, to enhance this model's interpretability, SHapley Additive exPlanations (SHAP) values were calculated, providing valuable insights into the contribution of key variables to the model's output.

## I. INTRODUCTION

Thyroid cancer is a type of cancer that begins in the tissues of the thyroid gland, a small organ in the front, lower neck region that is essential for controlling metabolism through the hormone synthesis process [1,2]. Initially, when thyroid cancer is first developing, it may show no or few symptoms [3]. However, when symptoms appear, a clear lump or swelling in the neck, changes in voice, difficulty in swallowing, neck pain, or a recurrent cough that isn't related to a cold can be observed [4, 5]. Owing to its mild presentation, the disease may not be identified until it becomes severe or may be discovered by chance while imaging for another medical condition.

Thyroid cancer is the most common and widespread, making up over 95% of all endocrine-related cancers [4]. The prevalence of thyroid cancer has been rising significantly over the last few decades, partly due to improved imaging and diagnostic techniques. Although thyroid cancer is a rare cancer type compared to other cancers [6], it is still a significant public health issue, due to its increasing discovery rates and the fact that it frequently requires long-term medical care. However, nearly 90% of thyroid cancers are identified as differentiated thyroid cancers [7], making it the most common type of thyroid cancer, which comprises two histological subtypes: papillary thyroid carcinoma (PTC) and follicular thyroid carcinoma (FTC) [8].

DTC patients often record a high survival rate, which is more than 85% [9], indicating favourable results. Although the survival rate is high, its recurrence is still a major clinical issue, which occurs in nearly 10-30% of DTC patients [10]. This leads to major challenges in patient care, including lifetime surveillance, multiple surgeries, radioactive iodine therapy, etc. Therefore, accurate classification and prediction of DTC recurrence are essential to improve and enhance overall patient care.

Conventional prognostic tools, especially those introduced by the American Thyroid Association (ATA), facilitate a systematic way to assess the probability of recurrence of DTC. However, as data complexity and dimensionality increase, more flexible analytical methods are needed to classify DTC recurrence more accurately. As a result, machine learning models have gained attention due to their ability to correctly classify the DTC recurrence, particularly using high-dimensional clinical and pathological data, thereby identifying complex relationships.

Therefore, initially, this study employs several ML models using the complete set of variables in order to classify DTC recurrence. To reduce the potential redundancy as well as the noise arising from all the variables, feature selection is carried out, and then the same ML models are applied to the reduced dataset to evaluate the improvement in the models' performance. With the use of this two-step process, this study aims to assess how feature selection affects model behaviour systematically. Although these ML models perform well, they provide point estimates rather than giving the confidence or uncertainty of the model's output. To address this challenge, a BNN is employed in this study, as it is capable of handling nonlinear and complex relationships while quantifying uncertainties by placing probabilistic weight distributions on model parameters. In addition, this study aims to examine how different prior distributions affect this model's performance and uncertainty estimations both before and after feature selection. Following that, the best-performing model was subjected to SHAP analysis in order to interpret feature contributions to the model's prediction, thereby reducing the black-box nature of the classification models. This post hoc interpretability step supports clinical transparency and increased confidence in AI-assisted categorisation by offering insightful information about model decision-making. Therefore, this study tries to provide a comprehensive framework to classify differentiated thyroid cancer recurrence by integrating feature selection, machine learning models, uncertainty-aware Bayesian modelling, and SHAP analysis together. Thereby, this study contributes to the advancements in the field of differentiated thyroid cancer data analysis by integrating robust, accurate, and interpretable classification methods.

The remainder of this paper is structured as follows: Section II provides the existing literature related to different classification techniques related to DTC recurrence. The data and materials used in this study and the methods carried out are outlined in Section III, while Section IV thoroughly discusses the results obtained in this study. Finally, Section IV concludes this paper.

## II. LITERATURE REVIEW

In recent years, machine learning techniques have been widely used in DTC recurrence classification utilising clinical and pathological data. These ML methods are capable of capturing nonlinear and complex relationships, which act as an alternative method to standard prognostic systems. Multiple studies [11,12,13, 14,15,16,17,18] have thoroughly examined the performance of classical machine learning models such as LR, k-Nearest Neighbours (KNN), SVM, Naïve Bayes (NB), and Stochastic Gradient Descent (SGD), revealing their higher accuracies ranging from 92% to 98%. In addition, a study [12] further evaluated the impact of various kernel functions, including linear, polynomial, and radial basis function (RBF) on SVMs and observed that the SVM with the RBF function achieved an accuracy of 94%, while balancing recall and F1-score effectively. Furthermore, the studies mentioned above and also the research project conducted by Şeyma Yaşar [19] explored tree-based models including Random Forest (RF), Decision Trees (DT), and Extra Tree Classifier (ETC), and identified that these models consistently showed outstanding results in DTC recurrence classification. Notably, the RF model exhibited strong performance along with exceptional sensitivity and specificity scores. However, despite the higher accuracies, these models are easily prone to overfitting, especially with small and imbalanced datasets, highlighting their limitations in applicability. Furthermore, these models need careful hyperparameter tuning to ensure their generalizability.

In addition, previous studies have also applied more sophisticated ensemble methods, such as Adaptive Boosting (Adaboost), Gradient Boosting (GB), Extreme Gradient Boosting (XGB), Light Gradient Boosting (LGB), and Catboost [11,12,13,14,15,16,19,20] models. These models obtained accuracies exceeding 85% by combining weak learners, forming strong classifiers, and improving the overall performance. Moreover, stacking ensemble models often performs well by aggregating multiple base learners. Few studies [14,18] have employed stacking ensemble models, demonstrating their superior performance, achieving higher accuracy. However, these ensemble methods are often computationally extensive and challenging to train, which indicates their implementation complexity. To improve the predictive performance further, the authors have implemented several hybrid models by integrating different machine learning techniques. Hybrid models such as RF-LR, SGD-LR, and SVM-LR, which combined multiple machine learning models, acquired nearly 97% accuracy, showing better performance [14].

Furthermore, to capture more intricate patterns, building upon the foundations of machine learning techniques, several authors [14,16,17,21] have explored some deep learning models, including Artificial Neural Networks (ANN) and feedforward Neural Networks (FNN), for DTC recurrence classification.

These studies reported that deep learning models were capable of achieving accuracies ranging from 96% to 98% highlighting their strength in modelling complex and nonlinear relationships. However, these models often require a large amount of data to train the model effectively, making them less suitable for small datasets.

Deviating from traditional machine learning and deep learning models, a study [14] explored the possibility of applying clustering techniques for DTC recurrence classification. The authors employed unsupervised clustering techniques such as k-means, hierarchical clustering, and Louvain clustering to uncover underlying patient subgroups. Notably, k-means and hierarchical clustering identified two clusters achieving higher silhouette scores and demonstrating that the dataset could be divided clinically into two major subgroups. However, these clustering techniques are highly sensitive to distance metrics and hyperparameters, which may lead to meaningless clusters if not carefully tuned or validated with expert opinions.

While these machine learning and deep learning models exhibit high performance, achieving remarkable accuracies, they often face difficulties with transparency and interpretability, making them less suitable for clinical decision making. Therefore, to overcome this black-box nature, a few studies have examined different post hoc explanation approaches, while some authors have employed significant explainable models to classify DTC recurrence data. Of these model explanation techniques, SHAP and Local Interpretable Model-agnostic Explanations (LIME) have been applied in a few studies to interpret how the features impact several ML models' outputs [13,20,21]. The authors highlighted the importance of SHAP and LIME values in identifying significant features as they provide notable decision-making information at both global and local levels. In contrast, A. K. Arslan and C. Çolak [22] implemented two explainable models, including Explainable Boosting Machines (EBM) and Fast Interpretable Greedy-Tree Sums (FIGS), to classify recurrence data. EBM exhibited a test accuracy of 96.10% while identifying key predictors. However, these model explanation techniques, as well as explainable models, should be handled carefully, as incorrect interpretations could lead to misleading information. Furthermore, few studies [11,12,14,16,17,19,20,22] have emphasised the role of feature selection in producing reliable and straightforward models for classification. Most common feature selection techniques, such as recursive feature elimination, feature importance analysis using RF and DT, Chi-square test, distance-based correlation, filter-based feature selection, and correlation heatmap, have been used to reduce the dimensionality and enhance the model stability by identifying the most significant features. However, there is a significant risk of discarding relevant features while implementing these techniques if the thresholds are not calibrated well.

Although ML, deep learning, and explainable models exhibit strong predictive performance, they often face challenges with uncertainty quantification. This limits the confidence not only in decision-making but also in model predictions, which is crucial for clinical applications. Furthermore, traditional ML models often produce a point estimate without specifying the degree of certainty in their output. This could lead to serious misclassifications, which could result in either excessive or insufficient treatment, making it dangerous for patient care. However, to overcome these challenges, Bayesian Neural Networks have emerged as a more flexible alternative approach, which facilitates the incorporation of prior distributions while quantifying uncertainties by producing probabilistic distributions. Rather than using fixed weights, BNNs learn distributions over weights while acquiring the strengths of the neural networks. Despite these advantages, no study has employed the BNN to classify DTC recurrence up to now, representing a significant research gap. Although some studies have already compared ML models, no studies have compared them with BNN in terms of accuracy, interpretability, and uncertainty. Considering these gaps, this study tries to provide a comprehensive comparison of ML models with BNN while varying prior distributions, offering a systematic framework for DTC recurrence classification. To further enhance the understanding of the model's predictions, this study provides a detailed explanation of the BNN's output using SHAP values, which have not yet been discussed in any previous literature. By filling these gaps, this study intends to offer a clinically usable model that facilitates confidence in data-driven decision-making in the management of DTC recurrence by combining both explainability and accuracy.

## III. METHODOLOGY
### A. Data
This study used a dataset related to DTC, which was obtained from an online data repository [31]. The dataset consisted of 16 variables representing the clinical and pathological features of 383 patients in a retrospective cohort who were followed up for at least 10 years within a study span of 15 years. The features of this dataset are the age of the patient in years, gender, present smoking status, whether the patient previously smoked or not (Hx Smoking), whether the patient received radiotherapy to the head and neck region or not (Hx Radiotherapy), thyroid function categories, Goiter type, adenopathy location, pathological subtype of cancer, focality, risk type, tumor stage (T), node status (N), metastasis status (M), cancer stage (stage), and response to initial treatment. The dependent variable of the dataset is whether the thyroid cancer recurred or not.

Initially, descriptive statistics were calculated in order to understand the characteristics of the dataset. Then, data preprocessing techniques such as standardisation, label encoding, and ordinal encoding were applied to the data. After that, this pre-processed data was divided into two sets: the training set (which contained 80% of the total dataset) and the testing set (which contained 20% of the total dataset). However, as a class imbalance was observed in this training set, the Synthetic Minority Oversampling Technique (SMOTE) was applied to achieve a more balanced distribution. Following that, firstly, 11 machine learning models were applied to the dataset containing all the variables in order to assess the baseline performance before conducting any feature selection.

B. Machine Learning Models

*1) Support Vector Machines:* SVM is one of the most commonly used machine learning models for classification as well as regression [23]. This is a supervised machine learning technique that can be used to classify both linear and nonlinear data by finding the optimal hyperplane that separates data into classes, using support vectors and margins [24].

*2) Random Forest:* RF is an ensemble model that fits multiple decision trees to arbitrary subsets of data, where each node in a decision tree applies a condition on a feature, splitting the data based on that condition [25]. Predictions from each decision tree are then combined through majority voting or averaging to increase the accuracy of the final prediction [26].

*3) K-nearest Neighbours:* Using a distance measure, KNN tries to find a test sample's k-nearest neighbours in the training set and classifies it according to the KNNs' majority vote. KNN's success is closely related to the neighbourhood that is searched for every sample, namely the hyperparameter k and the distance metric that is utilised [27].

*4) Decision Tree:* With a tree-like structure, DT tries to learn the basic decision rules derived from the features of the dataset and aims to build a model that predicts the output value. In order to draw conclusions about the output value, a tree is constructed by deciding to divide the source set into subsets. When all potential options and outcomes have been taken into account, the process comes to an end, producing a structure resembling a tree where the original decision is reached by following a logical chain of decisions [28].

*5) Logistic Regression:* The relationship between a group of independent factors and a categorical dependent variable can be studied using the LR model. Explanatory variables might be binary, continuous, discrete, or a combination of these, while the dependent variable is often dichotomous. To begin, it is necessary to prove that the collection of predictors and the outcome are related. Once a smaller group of predictor variables has been identified, the LR model can be applied to predict new outcomes [29].

*6) Gradient Boosting:* Similar to RF, the gradient boosting method makes predictions using the outputs of several decision trees, and then gradient trees use these decision trees as weak learners. In contrast to RF, the GB model builds one tree at a time, correcting each error of the preceding trees before it [30].

*7) Extreme Gradient Boosting:* As an alternative technique for predicting a response variable given specific covariates, XGB was suggested, and this uses a gradient boosting framework, which is an ensemble method based on decision trees. The fundamental idea behind this approach is to construct classification or regression trees one at a time, then train the next model using the residuals from the prior tree. This technique can use values from previously trained trees in the training process to get superior results, while it can prevent overfitting by employing a pruning process [31].

*8) Light Gradient Boosting:* LGB is also based on a gradient boosting framework, which trains several weak learners in succession while attempting to fix others' errors. A collection of these weak learners that come together to become a strong learner makes up the final model by reducing the loss function, which is usually the total of the losses for every case in the dataset [32].

*9) Catboost:* Catboost is a machine learning model based on symmetric decision trees, and by preprocessing features and creating additional numerical features, this model facilitates the usage of several feature categories in contrast to XGB and LGB. As a result, feature normalisation is not required in the catboost method, and it increases the dimension and utilisation of features at the model level by accounting for the association between attributes [33].

*10) Multi-Layer Perceptron:* MLP is a neural network that is made up of several layers, such as an input layer, hidden layers, and an output layer. In addition to receiving input from every unit in the layer before it, each unit transmits its output to every other unit in the layer beneath. There is a certain weight assigned to each of these links, which shows how the unit affects the unit's response in the layer afterwards. The strength of these connections between the units and the input determines a multilayer perceptron's output [34].

*11) Naïve Bayes:* NB is a popular classifier that is based on Bayes' theorem, which makes the assumption that a feature's impact on a class is not influenced by other features. NB returns the conditional probability of a specific target after computing the prior probabilities and likelihood for a given class label [35].

However, to optimise these models' parameters and evaluate their performances, a grid search with 2-fold, 5-fold, and 10-fold cross-validation was performed for each machine learning model, and then the model that produced the highest accuracy across these folds was selected for final testing. In addition to accuracy, several other evaluation metrics, such as precision, recall, and F1-score, were calculated for each of these machine learning models. To calculate these evaluation metrics, a confusion matrix, which is depicted in Table 1, was used.

Table 1. Confusion matrix

|  |  | Predicted Class | |
|---|---|---|---|
|  |  | Negative | Positive |
| Actual Class | Negative | True Negative (TN) | False Positive (FP) |
|  | Positive | False Negative (FN) | True Positive (TP) |

Using the above confusion matrix in Table 1, accuracy, precision, recall, and F1-score were calculated using Equations 1, 2, 3, and 4, respectively:

$$\text{Accuracy} = \frac{TP+TN}{TP+TN+FP+FN} \quad (1)$$

$$\text{Precision} = \frac{TP}{TP+FP} \quad (2)$$

$$\text{Recall} = \frac{TP}{TP+FN} \quad (3)$$

$$\text{F1-score} = \frac{2 \times \text{Precision} \times \text{Recall}}{\text{Precision} + \text{Recall}} \quad (4)$$

Following that, to reduce the model complexity and increase the model training efficiency, a feature selection was performed using the Boruta algorithm, thereby identifying the relevant features and effectively removing the redundant variables. The Boruta algorithm is a feature selection method and also an extension of the RF method, which contrasts the importance of original variables with that of their randomly generated shadow counterparts in order to identify the most relevant features [36].

After this feature selection process, the same machine learning models that are mentioned in Section B were applied to the reduced dataset containing only the selected variables. In addition, the same grid search and cross-validation procedure that was discussed earlier was performed with the reduced dataset to improve the models' performance. To assess these machine learning models, accuracy, precision, recall, and F1-score were calculated again.

However, it was observed that these machine learning models lack interpretability and fail to capture the uncertainty of the model's parameters. Therefore, to overcome these challenges, a simple Bayesian Neural Network was employed on both the full and reduced datasets, and its general overview is depicted in Figure 1.

*C. Bayesian Neural Network*

BNN is a type of artificial neural network that incorporates Bayesian inference during the training phase [37]. In contrast to typical ANNs, which use deterministic optimisation to learn a single set of optimal weights, BNNs regard the model parameter vector θ as random variables and try to infer their posterior distributions from the observed data while quantifying the uncertainties [38].

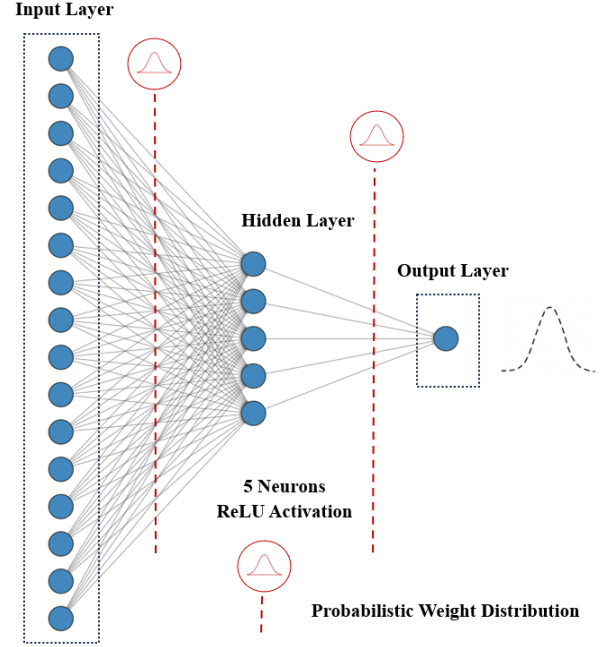

Figure 1. General overview of BNN

i.e., the BNN's goal is to estimate the full posterior distribution $p(\theta|D_x,D_y)$, which is derived using the Bayes' theorem, when,

$$p(\theta|D_x,D_y) = \frac{p(D_y|D_x,\theta)p(\theta)}{\int_\Theta p(D_y|D_x,\theta')p(\theta')d\theta'}. \quad (5)$$

Here, $D_x=\{x_1,x_2,\ldots,x_N\}$, is a matrix of variables and $D_y=\{y_1,y_2,\ldots,y_N\}$, is the corresponding output matrix when N indicates the number of observations. In addition, $p(D_y|D_x,\theta)$ and $p(\theta)$ represent the likelihood and prior distribution, respectively. Furthermore, the denominator is the marginal likelihood that integrates over the possible values of the parameter space and normalizes the posterior distribution. As this integral is difficult to compute in practice, and acts as a normalizing constant, the posterior distribution can be expressed as in Equation (6).

$$\text{posterior} \propto \text{likelihood} \times \text{prior} \quad (6)$$

However, in the binary classification context, the outputs $D_y=\{y_1,y_2,\ldots,y_N\}$, are binary labels. i.e. $y_i \in \{0,1\}$. Then the BNN considers the likelihood as a product of a Bernoulli distribution, which is expressed in Equation (7).

$$p(D_y|D_x,\theta) = \prod_{i=1}^{N} \hat{p}_i^{y_i}(1-\hat{p}_i)^{1-y_i} \quad (7)$$

In Equation (7), $\hat{p}_i$ indicates the network's output for the $i^{th}$ sample, representing the estimated probability that sample i is a member of class 1, where $\hat{p}_i = NN_\theta(x_i)$. The prior is placed over the network parameters θ. A common choice for the prior distribution is a diagonal Gaussian prior, which can be expressed as,

$$p(\theta) = N(0,\sigma^2 I) \quad (8)$$

where $\sigma^2$ is a scalar and I is the identity matrix [38].

By considering the Bayesian framework discussed above, this study employed a simple BNN for the dataset with all the variables and the dataset with selected variables using the Boruta algorithm. The model was built using the Pyro probabilistic programming language, and the model architecture was comprised of an input layer of dimension d which is corresponding to the number of covariates, a hidden layer with five units and ReLU activation, and a final output layer producing a single logit which is transformed to a Bernoulli likelihood for classification. Weights and biases were included in the network parameters θ, which were modelled as independent random variables, and several prior distributions were taken into consideration for the parameters in order to investigate how prior choice affects model performance and uncertainty quantification. These prior distributions included: N (0,1), N (0,10), Laplace (0,1), Cauchy (0,1), Cauchy (0,2.5), and Horseshoe (1). As these prior distributions are composed of varying sparsity-inducing properties as well as varying tail behaviour, they facilitated a comprehensive comparative analysis of the model performance.

Due to the practical difficulty of assessing the posterior distribution in closed form, the posterior distributions of these BNNs were estimated using the Markov Chain Monte Carlo (MCMC) method. This technique simulates direct draws from target posterior distributions without requiring the analytical solution of numerous integrals [39]. For this purpose, Hamiltonian Monte Carlo (HMC), which is a gradient-based MCMC technique that uses Hamiltonian dynamics to suggest effective transitions in the parameter space, was employed. The No-U-Turn Sampler (NUTS), an adaptive extension of HMC, was used to increase the sampling efficiency even further and eliminate the necessity of manually setting the expensive tuning runs [40].

Moreover, to evaluate each of these BNN models with varying prior distributions, the same evaluation metrics that were used to assess the machine learning models, including accuracy, precision, recall, and F1-score, were computed. Furthermore, epistemic uncertainty, which arises from limited data reflecting uncertainty in the model parameters, and aleatoric uncertainty, which captures the inherent noise in the data, were estimated for each BNN [41].

After carefully evaluating each machine learning and BNN model employed before and after feature selection, the model that produced the highest accuracy was selected as the best-performing model. Then, to assess how each feature of the dataset contributed to this model's predictions, SHAP values were calculated.

*D. Explainable Artificial Intelligence with SHAP*

When employing an ML model, it is crucial to identify the most significant and influential variables while understanding how they contribute to model predictions. However, this is limited in traditional ML models, which do not show how the features affect the model's predictions; instead, they provide how they produce the model's output [42]. To reduce the black box nature of these ML models, SHAP has been introduced by integrating the concepts of game theory, forming part of the explainable artificial intelligence (XAI), thereby providing globally accurate and consistent explanations of feature importance and how they influence the model's predictions [43,44]. SHAP analysis allocates a specific contribution to each feature by treating each variable as a contributor in order to quantify the influence of the features on the predictions [45]. This not only quantifies the overall effect but also computes both the negative and positive influences of features on the output of the model [46]. By employing an Explanatory Model (EM), SHAP assesses each variable's contribution to the EM as in Equation 9, where the input data is represented as $X_i = \{x_1, x_2, \ldots, x_n\}^t$ [47].

$$EM = \varphi_0 + \sum_{i=1}^{n} \varphi_i t_i \quad (9)$$

where,

$$\varphi_i(ML, x) = \sum_{t \subseteq x}^{n} \frac{|t|!(n-|t|-1)!}{n!} [ML(t) - ML(t\setminus i)] \quad (10)$$

Here, $i^{th}$ input variable is represented by $t_i$ when there are n observations. $\varphi_0$ denotes the model output in the case where all simplified input features are inactive or set to their baseline values, and $\varphi_i$ indicates the $i^{th}$ variable's contribution to the original prediction ML model, while t corresponds to the fraction of the model's features that excludes the feature value i.

## IV.  RESULTS AND DISCUSSION

This dataset consisted of data related to 383 patients who participated in a retrospective cohort study on differentiated thyroid cancer recurrence. It contained 16 variables, including one numerical variable (the patient's age) and 15 categorical variables representing clinical and pathological features of patients. Figure 2 depicts the age distribution of the patients at diagnosis, and Table 2 describes the descriptive statistics calculated for patients' age.

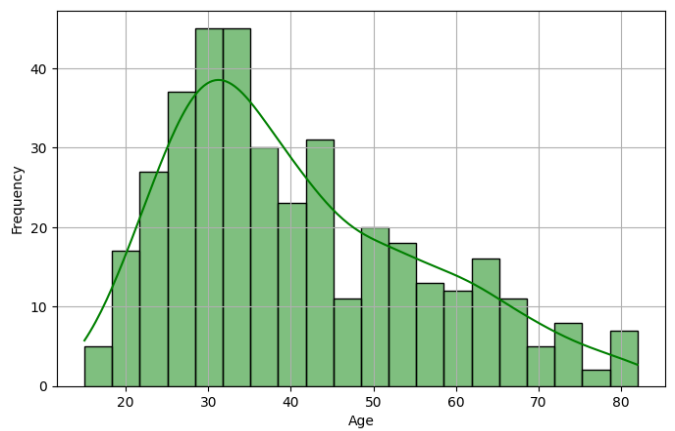

Figure 2.  Age distribution of the patients at diagnosis

Table 2. Descriptive statistics for the age variable

| Minimum | Maximum | Mean | Standard Deviation |
|---|---|---|---|
| 15 | 82 | 40.867 | 15.134 |

According to Figure 2, it is clear that the age distribution is right-skewed, indicating that the majority of the patients are younger, while fewer patients are older. However, patients' ages ranged from 15 to 82 years, while the peak of the distribution is observed in the 30-35 years range, indicating that most of the patients are within this range. In addition, the mean age of patients is 40.867 years ($\cong$ 41 years) while the standard deviation is 15.134 years. To gain a clear understanding of the dataset, the baseline characteristics of the patients by recurrence status are described in Table 3.

Table 3. Baseline characteristics of the patients by recurrence status

| Features | Recurred (%) | Not Recurred (%) |
|---|---|---|
| Age | 108 (28.20%) | 275 (71.80%) |
| Gender | | |
|     Male | 42 (10.97%) | 29 (7.57%) |
|     Female | 66 (17.23%) | 246 (64.23%) |
| Smoking Status | | |
|     Smoking | 33 (8.62%) | 16 (4.18%) |
|     Not Smoking | 75 (19.58%) | 259 (67.62%) |
| Past Smoking Status | | |
|     Smoked | 14 (3.66%) | 14 (3.66%) |
|     Never Smoked | 94 (24.54%) | 261 (68.15%) |
| Radiotherapy to the Head and Neck Region | | |
|     Received | 6 (1.57%) | 1 (0.26%) |
|     Not Received | 102 (26.63%) | 274 (71.54%) |
| Thyroid Function Categories | | |
|     Clinical Hyperthyroidism | 3 (0.783%) | 17 (4.439%) |
|     Clinical Hypothyroidism | 2 (0.522%) | 10 (2.611%) |
|     Euthyroid | 98 (25.857%) | 234 (61.097%) |
|     Subclinical Hyperthyroidism | 0 (0.000%) | 5 (1.305%) |
|     Subclinical Hypothyroidism | 5 (1.305%) | 9 (2.350%) |
| Goiter Type | | |
|     Diffuse | 0 (0.000%) | 7 (1.83%) |
|     Multinodular | 52 (13.58%) | 88 (22.98%) |
|     Normal | 2 (0.52%) | 5 (1.31%) |
|     Single Nodular Left | 26 (6.79%) | 63 (16.45%) |
|     Single Nodular Right | 28 (7.31%) | 112 (29.24%) |
| Adenopathy Location | | |
|     Bilateral | 27 (7.050%) | 5 (1.305%) |
|     Extensive | 7 (1.828%) | 0 (0.000%) |
|     Left | 12 (3.133%) | 5 (1.305%) |
|     No Adenopathy | 30 (7.833%) | 247 (64.491%) |
|     Posterior | 2 (0.522%) | 0 (0.000%) |
|     Right | 30 (7.833%) | 18 (4.700%) |
| Pathological Subtype of Cancer | | |
|     Follicular | 12 (3.13%) | 16 (4.18%) |
|     Hurthel cell | 6 (1.57%) | 14 (3.66%) |
|     Micropapillary | 0 (0.00%) | 48 (12.53%) |
|     Papillary | 90 (23.50%) | 197 (51.44%) |
| Focality | | |
|     Multi Focal | 70 (18.28%) | 66 (17.23%) |
|     Uni Focal | 38 (9.92%) | 209 (54.57%) |
| Risk Type | | |
|     High | 32 (8.36%) | 0 (0.00%) |
|     Intermediate | 64 (16.71%) | 38 (9.92%) |
|     Low | 12 (3.13%) | 237 (61.88) |
| Tumor Stage | | |
|     T1a | 1 (0.261%) | 48 (12.533%) |
|     T1b | 5 (1.305%) | 38 (9.922%) |
|     T2 | 20 (5.222%) | 131 (34.204%) |
|     T3a | 41 (10.705%) | 55 (14.360%) |
|     T3b | 14 (3.655%) | 2 (0.522%) |
|     T4a | 9 (4.961%) | 1 (0.261%) |
|     T4b | 8 (2.089%) | 0 (0.000%) |
| Node Status | | |
|     N0 | 27 (7.05%) | 241 (62.92%) |
|     N1a | 10 (2.61%) | 12 (3.13%) |
|     N1b | 71 (18.54%) | 22 (5.74%) |
| Metastasis Status | | |
|     M0 | 90 (23.50%) | 275 (71.80%) |
|     M1 | 18 (4.70%) | 0 (0.00%) |
| Cancer Stage | | |
|     I | 65 (16.971%) | 268 (69.974%) |
|     II | 25 (6.527%) | 7 (1.828%) |
|     III | 4 (1.044%) | 0 (0.000%) |
|     IVA | 3 (0.783%) | 0 (0.000%) |
|     IVB | 11 (2.872%) | 0 (0.000%) |
| Response to Initial Treatment | | |
|     Biochemical Incomplete | 11 (2.87%) | 12 (3.13%) |
|     Excellent | 1 (0.26%) | 207 (54.05%) |
|     Indeterminate | 7 (1.83%) | 54 (14.10%) |
|     Structural Incomplete | 89 (23.24%) | 2 (0.52%) |

When considering Table 3, it is clear that 28.20% of the total patients experienced thyroid cancer recurrence, while 71.80% did not experience it. Furthermore, this cancer mostly recurred in males, current smokers, and patients who received radiotherapy to the neck and head. In terms of thyroid function, euthyroidism was common among the non-recurrence group, while multinodular goiter type was frequently observed among the recurrence patients. When considering adenopathy location, it was observed that either adenopathy on the right side or no adenopathy was present in the patients with recurrence status, whereas the majority of the non-recurrence group had no adenopathy. According to Table 3, the papillary cancer subtype was common among the recurrence participants, while the micropapillary subtype was not observed in the same group. In addition, the individuals with multifocal tumors with T3a tumor stage, N1b node status, M1 metastasis status, and high cancer stages such as (III, IVA, IVB) had a higher likelihood of DTC recurrence. Moreover, patients who experienced cancer recurrence showed poor response to initial treatments, while the participants with non-recurrence status responded excellently to the treatments.

After calculating the dataset's basic statistics, traditional ML models were first applied to the training dataset containing all the variables. These models were trained using 2-fold, 5-fold, and 10-fold cross-validation and tested using the test dataset. Evaluation metrics, including accuracy, precision, recall, and F1-score, were calculated for each of these cross-validation settings, and Table 4 depicts the highest metrics obtained among the cross-validation results.

Table 4. Evaluation metrics for machine learning models before the feature selection

| Model | Accuracy | Precision | Recall | F1-Score |
|---|---|---|---|---|
| SVM | 0.9481 | 0.8947 | 0.8947 | 0.8947 |
| RF | 0.9221 | 0.9333 | 0.7368 | 0.8235 |
| KNN | 0.8701 | 0.68 | 0.8947 | 0.7727 |
| DT | 0.8961 | 0.8236 | 0.7368 | 0.7778 |
| LR | 0.9359 | 0.8888 | 0.8421 | 0.8649 |
| GB | 0.9221 | 0.7826 | 0.9474 | 0.8571 |
| XGB | 0.8831 | 0.7083 | 0.8947 | 0.7907 |
| LGB | 0.8571 | 0.6426 | 0.9474 | 0.7659 |
| Catboost | 0.9091 | 0.7727 | 0.8947 | 0.8293 |
| MLP | 0.8684 | 0.6538 | 0.9444 | 0.7727 |
| NB | 0.8816 | 0.7143 | 0.8333 | 0.7692 |

According to Table 4, it is clear that all models have recorded higher accuracies, with each model obtaining above 0.85. However, overall, SVM performed well, achieving a higher accuracy of 0.9481 with a 0.8947 precision, 0.8947 recall, and 0.8947 F1 score, making it suitable for classification. Furthermore, the LR model also performed better, recording the second-highest accuracy. Despite the better performance, it is necessary to assess the confusion matrix obtained for the SVM model. Therefore, the confusion matrix of the SVM is demonstrated in Figure 3.

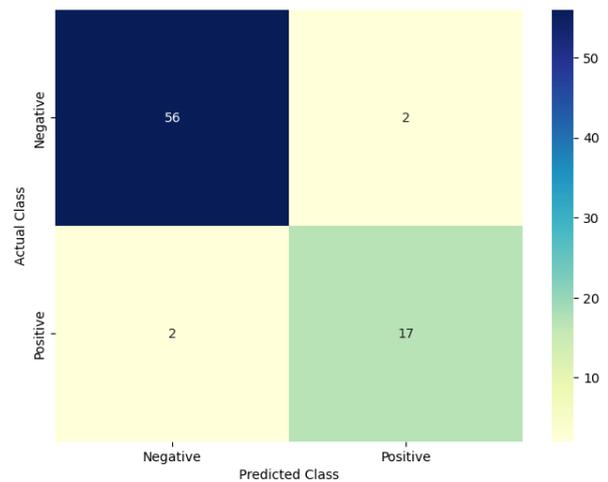

Figure 3. Confusion matrix for the SVM model before feature selection

When considering Figure 3, it is evident that the SVM model was able to make correct predictions with a higher degree of accuracy and fewer misclassifications. Although the SVM model performed well with the complete set of variables, systematically identifying the most relevant features that have a greater predictive power is crucial to reducing the model's complexity and providing robust results as well as interpretations. Therefore, a feature selection was performed using the Boruta algorithm, and the most influential features were obtained, as shown in Figure 4.

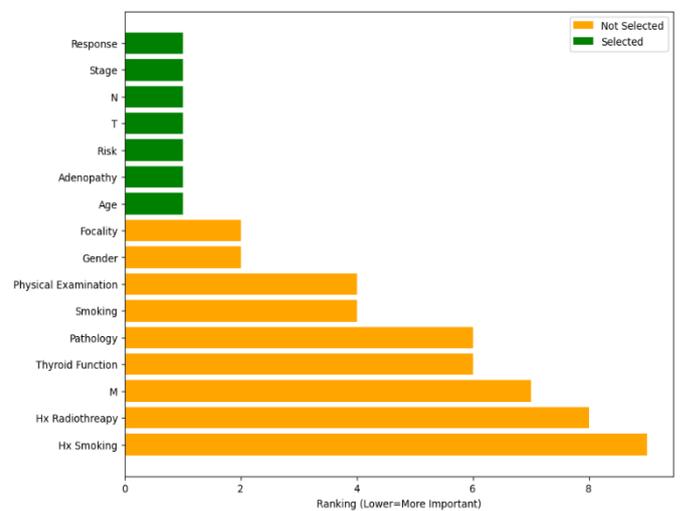

Figure 4. Feature importance by the Boruta algorithm

According to Figure 4, response to initial treatment, cancer stage, node status, tumor stage, risk type, adenopathy location, and age are the most relevant features selected by the Boruta algorithm. Considering this, the same machine learning models were applied for the selected features and trained using 2-fold, 5-fold, and 10-fold cross-validation. These models were tested using test data, and accuracy, precision, recall, and F1-score were calculated for three cross-validation settings, and Table 5 indicates the highest metrics obtained across validation.

Table 5. Evaluation metrics for machine learning models after the feature selection

| Model | Accuracy | Precision | Recall | F1-Score |
|---|---|---|---|---|
| SVM | 0.9610 | 0.9000 | 0.9474 | 0.9231 |
| RF | 0.9481 | 0.8571 | 0.9474 | 0.9 |
| KNN | 0.8831 | 0.7083 | 0.8947 | 0.7907 |
| DT | 0.9091 | 0.8333 | 0.7894 | 0.8108 |
| LR | 0.9611 | 0.9444 | 0.8947 | 0.9189 |
| GB | 0.9351 | 0.8823 | 0.7894 | 0.8833 |
| XGB | 0.8961 | 0.7619 | 0.8421 | 0.8 |
| LGB | 0.8701 | 0.68 | 0.8947 | 0.7727 |
| Catboost | 0.9351 | 0.85 | 0.8947 | 0.8718 |
| MLP | 0.8701 | 0.6667 | 0.9474 | 0.7826 |
| NB | 0.8961 | 0.7391 | 0.8947 | 0.8095 |

According to Table 5, the LR model outperformed the other models, achieving an accuracy of 0.9611, a precision of 0.9444, a recall of 0.8947, and an F1-score of 0.9189. Followed by that, the SVM model indicates an accuracy of 0.9610, which is slightly lower than that of the LR model. It is noted that each of the models has achieved a higher accuracy after feature selection, as shown in Figure 5, indicating that removing irrelevant features improves the model performance. Therefore, the LR model is identified as the best-performing model across all the machine learning models after feature selection, and its confusion matrix is denoted in Figure 6.

As shown in Figure 6, the LR model performs well with fewer misclassifications, achieving higher accuracy and indicating its capability in the classification of thyroid cancer recurrence. Although these traditional machine learning models performed well before and after feature selection, they lack uncertainty quantification, offering limited interpretability, particularly with small datasets. Therefore, to overcome these challenges with uncertainty quantification in traditional machine learning models, this study implemented a Bayesian Neural Network on the dataset before and after feature selection. The prior distributions of this model were varied from the standard normal distribution to the horseshoe prior, and accuracy, precision, recall, and F1-score for each model were calculated as depicted in Table 6.

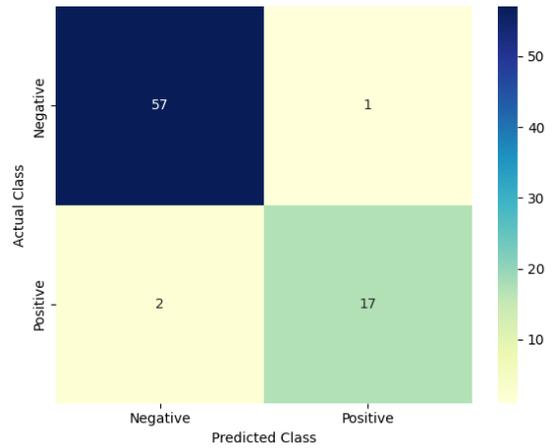

Figure 6. Confusion matrix for the LR model after feature selection

Table 6. Evaluation metrics of the BNN model before feature selection

| Model | Prior | Accuracy | Precision | Recall | F1-Score |
|---|---|---|---|---|---|
| Before feature selection | Normal (0,1) | 0.9221 | 0.8823 | 0.7894 | 0.8333 |
| | Normal (0,10) | 0.9740 | 0.9474 | 0.9474 | 0.9474 |
| | Laplace (0,1) | 0.9221 | 0.8095 | 0.8947 | 0.8500 |
| | Cauchy (0,1) | 0.9091 | 0.8750 | 0.7368 | 0.8000 |
| | Cauchy (0,2.5) | 0.9351 | 0.9375 | 0.7894 | 0.8571 |
| | Horseshoe (1) | 0.8961 | 0.7200 | 0.9474 | 0.8182 |
| After feature selection | Normal (0,1) | 0.9359 | 0.8888 | 0.8421 | 0.8649 |
| | Normal (0,10) | 0.9870 | 1.0000 | 0.9474 | 0.9729 |
| | Laplace (0,1) | 0.9481 | 0.8947 | 0.8947 | 0.8947 |
| | Cauchy (0,1) | 0.9351 | 0.8182 | 0.9474 | 0.8780 |
| | Cauchy (0,2.5) | 0.9481 | 0.9412 | 0.8421 | 0.8889 |
| | Horseshoe (1) | 0.9091 | 0.8750 | 0.7368 | 0.8000 |

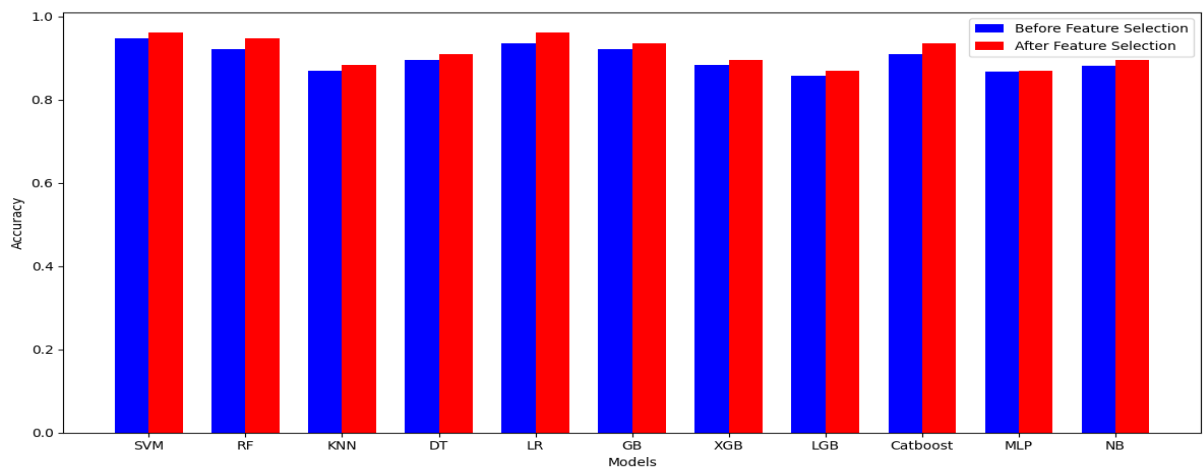

Figure 5. Model accuracies before and after feature selection

According to Table 6, it is noted that the BNN model with the Normal (0,10) prior distribution achieved the highest accuracy of 0.9740 with a precision of 0.9474, recall of 0.9474, and F1-score of 0.9474 across six prior distributions before feature selection. This surpassed the SVM model, which showed a higher accuracy across all the ML models employed before the feature selection. The confusion matrix of this BNN model, which is illustrated in Figure 7, also confirms that this model performed well with fewer misclassifications.

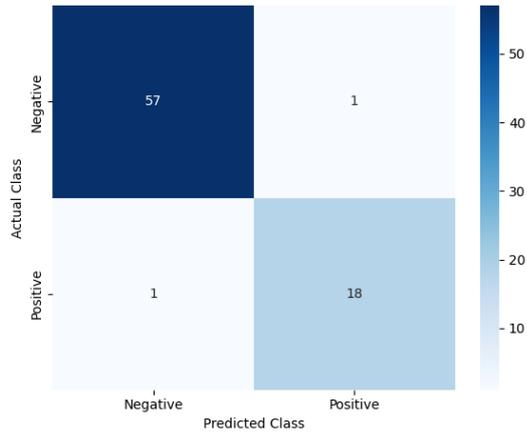

Figure 7. Confusion matrix for the BNN model with Normal (0,10) prior distribution before feature selection

Similarly, BNN with the same prior obtained a maximum accuracy of 0.9870 after feature selection, indicating the model's effectiveness in capturing complex relationships while quantifying uncertainties.

However, when comparing the accuracies of the BNN and machine learning models used in this study, the BNN model with Normal (0,10) prior outperformed all the other models both before and after the feature selection, which is illustrated in Figure 8. This indicates that the BNN model with a less restrictive prior enhances the predictive performance of the model by accurately estimating parameters, thereby highlighting the model's flexibility in parameter estimation through posterior distributions.

Therefore, the BNN model with Normal (0,10) prior distribution employed on the feature-selected dataset was chosen as the best-performing model for thyroid cancer recurrence classification over all the machine learning and Bayesian models discussed in this study. To further analyse, the confusion matrix of this model was generated and depicted in Figure 9.

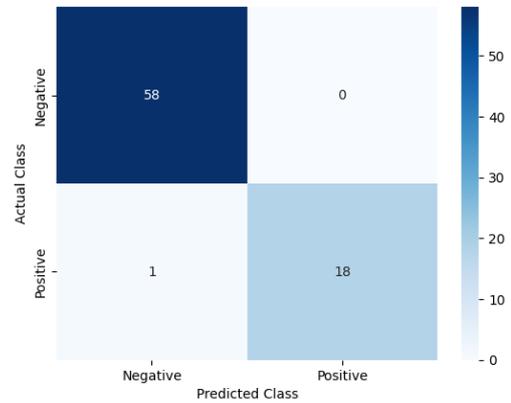

Figure 9. Confusion matrix of Bayesian Neural Network with N (0,10) prior distribution after feature selection

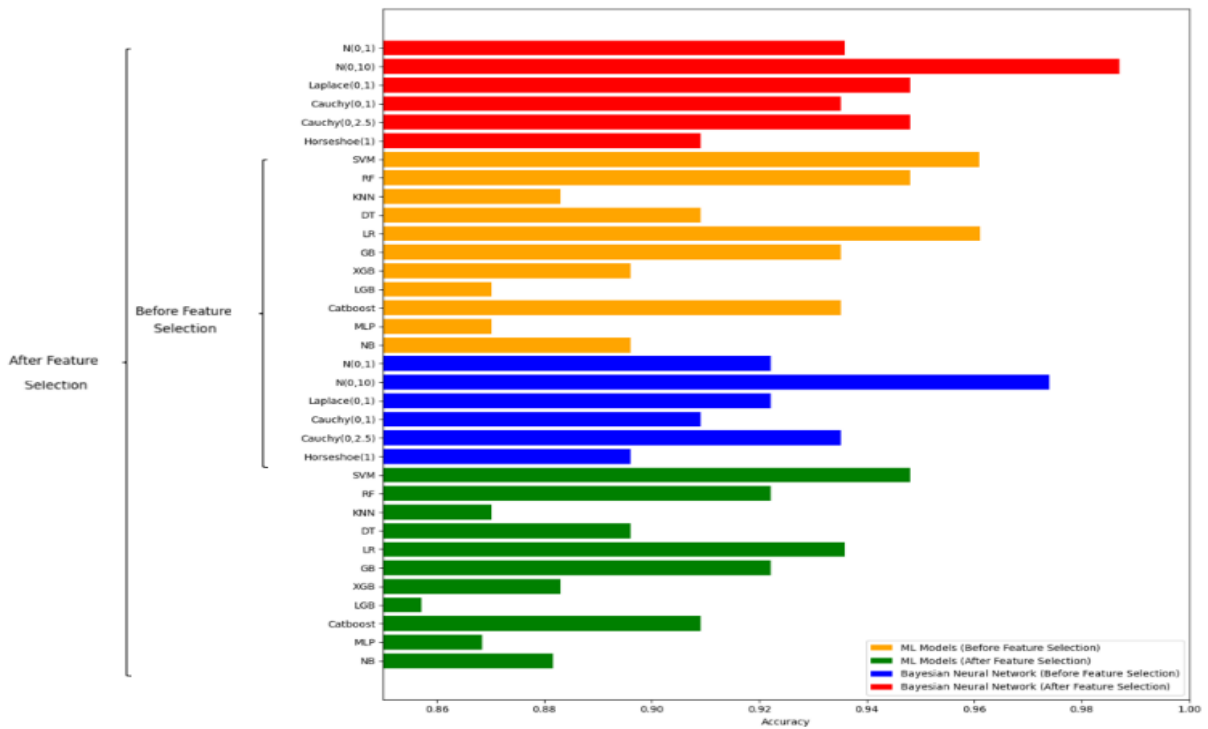

Figure 8. Accuracies of the machine learning and Bayesian models before and after feature selection

When considering Figure 9, it is evident that the BNN with Normal (0,10) prior performed excellently with only one misclassification while achieving a precision of 1.000. In Bayesian models, it is vital to visualise the epistemic uncertainties in order to identify the uncertainties in model predictions. In addition to epistemic uncertainties, it is equally crucial to visualise aleatoric uncertainties, which result from intrinsic noise or uncertainty in the data itself. Therefore, the posterior predictive mean with epistemic uncertainties as well as aleatoric uncertainties is depicted in Figures 10 and 11, respectively.

In these figures, the purple and blue lines represent the mean predicted probability for the test samples, and the shaded regions indicate the epistemic uncertainty and aleatoric uncertainty. In Figure 10, the narrow-shaded area highlights the higher confidence, indicating low epistemic uncertainty, where predictions are consistent across posterior samples, while the wider regions show lower confidence, demonstrating the model's high uncertainty. Similarly, the aleatoric uncertainties depicted in Figure 11 illustrate the uncertainty inherent in the data itself through the shaded areas. In addition, several test samples in both Figures 10 and 11 depicted higher probabilities, which are closer to 1 with less uncertainty. This shows the ability of the BNN model to accurately predict whether the thyroid cancer recurred.

Furthermore, these plots depict probabilities that fall within the intermediate range, where the majority of them are accompanied by wider uncertainty. This highlights the model's reduced confidence in predicting the thyroid recurrence class and indicates that these test samples should be handled carefully and that further evaluation is needed.

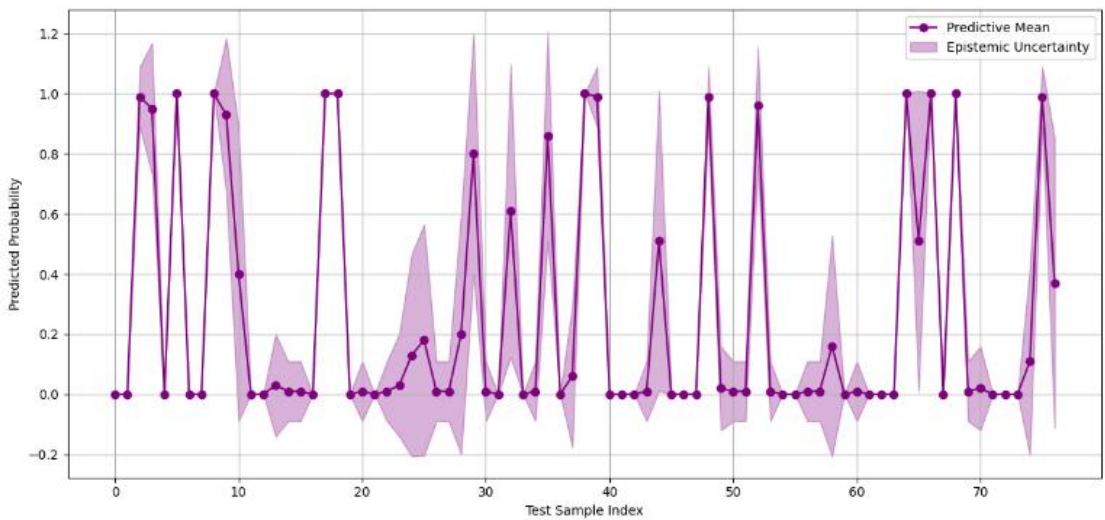

Figure 10. Posterior predictive mean with epistemic uncertainty

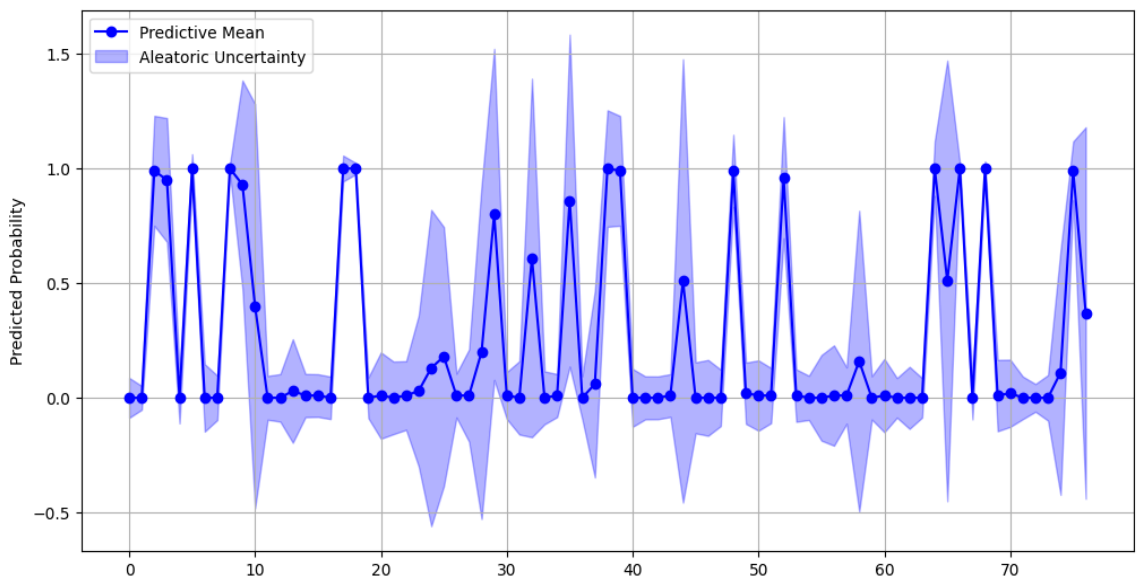

Figure 11. Posterior predictive mean with aleatoric uncertainty

However, while evaluating how confident the BNN model is in its predictions using epistemic and aleatoric uncertainty plots, it is crucial to assess why the model made these predictions, even if it performed well. Therefore, SHAP values were calculated for the selected BNN model to examine how each feature affects the final predictions, providing insights into the model's output. Initially, a global interpretation was carried out using SHAP values, in order to provide a brief understanding of the model's behaviour. Firstly, a summary bar plot of aggregated mean SHAP values was obtained as in Figure 12 to assess the overall importance of the features contributing to final predictions.

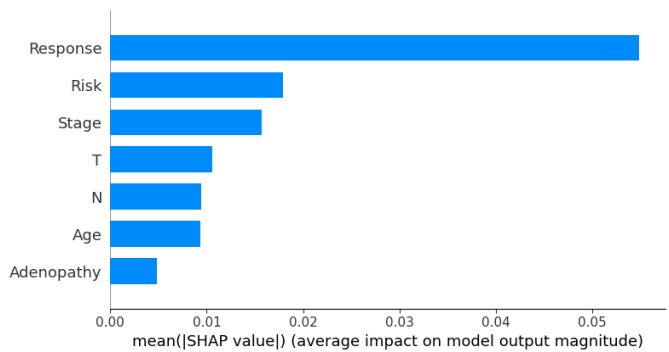

Figure 12. Bar plot of mean aggregated SHAP values for the BNN model after feature selection

According to Figure 12, it is observed that the response to initial treatment acts as the most influential factor among the seven variables in deciding whether the differentiated thyroid would recur or not. Followed by that, risk type and cancer stage contribute significantly to the final output, highlighting their role in cancer recurrence. Furthermore, tumor stage, node status, age, and adenopathy moderately affect the model's output. Although Figure 12 provided an overview of how each feature affects the final prediction, it is crucial to examine how each feature is varied across patients. Therefore, a SHAP beeswarm plot was obtained as in Figure 13, providing a global interpretation of the selected BNN model.

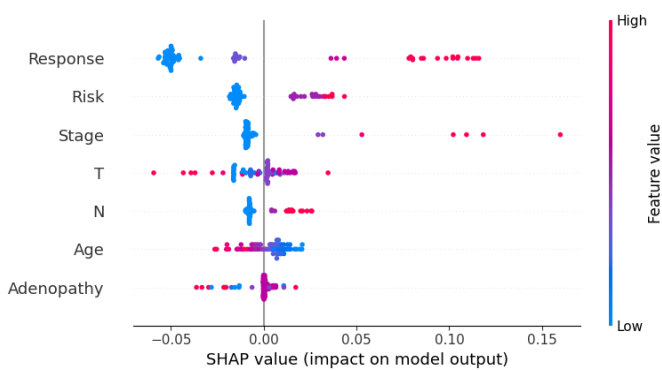

Figure 13. SHAP beeswarm plot for the selected BNN model

The beeswarm plot ranks the features according to their mean absolute SHAP value, where the features with a higher mean absolute SHAP value are located at the top, while the features with a lower value are at the bottom. Each dot in Figure 13 represents a patient, which indicates the magnitude as well as the direction of the SHAP value for a particular feature, where the instances with higher SHAP values are colored in red, while the cases with lower SHAP values are in blue. For example, response to initial treatment indicates a strong influence on the model's output, where a patient with red dots increases the prediction, showing a robust association with the recurrence of thyroid cancer. However, this demonstrates that the patients with higher severity status, such as structural incomplete responses to initial treatments, may have an increased recurrence of thyroid cancer. In contrast, low SHAP values indicate that the excellent response to initial treatments negatively affects the recurrence of thyroid cancer.

In addition, higher stages of risk type, cancer stage, and node status indicate positive SHAP values, suggesting that as the severity of these features increases, the probability of recurrence of thyroid cancer increases. However, when considering the tumor stage, it is observed that there are both red and purple plots, slightly placed on the right side of the plot, while some points are near zero, and a few are on the left side. This reveals that the larger tumors raise the risk of recurrence, while the effect of this feature is associated with other variables that need further clinical investigations.

Furthermore, the age variable shows a moderate impact on the model's predictions, where the younger age patients that are depicted in blue dots indicate a higher risk of thyroid cancer recurrence, while older patients have a lower risk or neutral effect on cancer recurrence. In addition, lower SHAP values for adenopathy demonstrate less impact on the model's output. However, this shows a mix of negative and positive contributions, highlighting that the effect of adenopathy is subtle and further investigations are required.

In addition to the SHAP global interpretation, this study focused on generating plots regarding SHAP local interpretations to provide a detailed explanation about the predictions for each instance in the dataset. Therefore, to assess how each of these features affects the predictions for each patient sequentially, a SHAP decision plot was obtained, as shown in Figure 14.

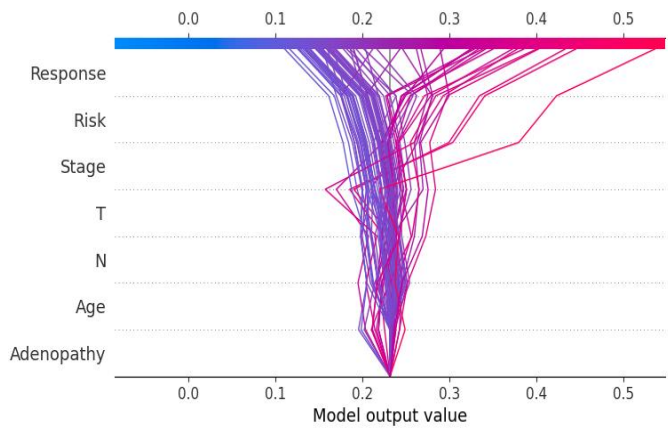

Figure 14. SHAP decision plot for the selected BNN model

When considering the decision plot in Figure 14, each prediction line starts from a common baseline value and then progressively shifts the direction of each line of all features, either increasing or decreasing the risk of thyroid recurrence. Response to initial treatment indicates a significant shift in the prediction, highlighting its effect in predicting the recurrence status. Notably, the poor response to initial treatment suggests an increased probability of cancer recurrence, which is evident in the right shift of the prediction lines in Figure 14. Similarly, risk type, cancer stage, and node status also exhibit a positive impact on the recurrence prediction, especially in patients with higher values. In addition, tumor stage and adenopathy display a deviated pattern depicting a decreased prediction risk, which suggests that these features are associated with other variables. However, age demonstrates a modest fluctuation where the prediction lines are observed in both directions. Furthermore, to examine how each of these features contributes to individual predictions, two waterfall plots for recurrence and non-recurrence classes were obtained for the same model as in Figures 15 and 16.

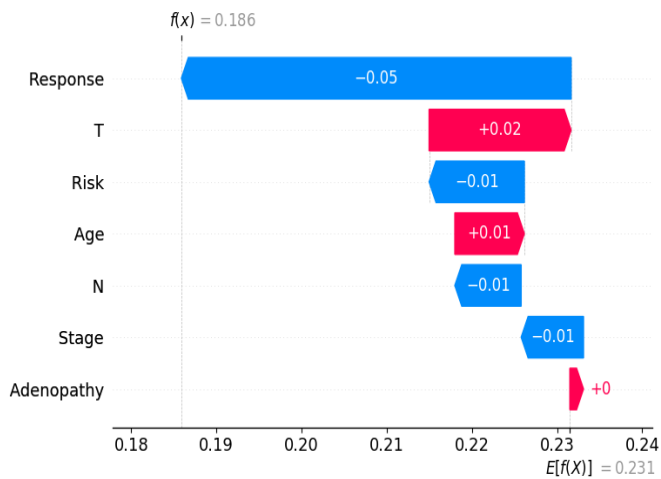

Figure 15. Waterfall plot for the non-recurrence class of the selected BNN model

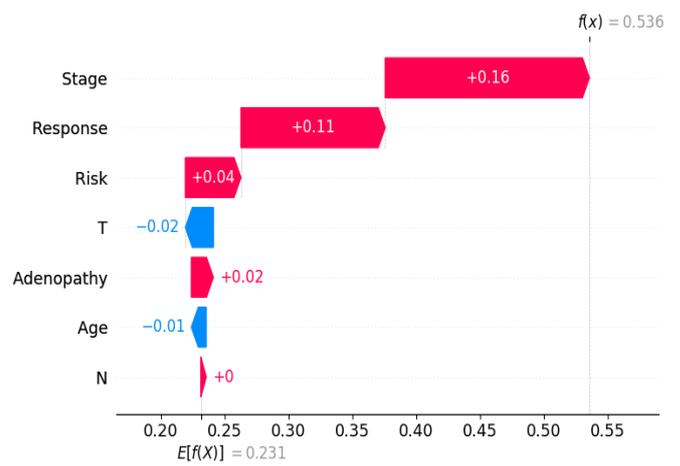

Figure 16. Waterfall plot for the recurrence class of the selected BNN model

Figures 15 and 16 provide a detailed explanation of the individual features that decide the thyroid cancer recurrence status of a patient, starting with an average recurrence probability of 0.231. According to Figure 15, the response feature shows a significant negative association with the non-recurrence prediction, indicating a SHAP value of -0.05. At the same time, it exhibits a SHAP value of +0.11, showing its contribution to recurrence prediction as in Figure 16. This shows that an excellent or indeterminate response to initial treatment notably reduces the risk of recurrence, while a poor response increases the risk. When considering Figure 15, it is observed that the lower levels of risk type, node status, and cancer stage reduce the likelihood of recurrence of thyroid cancer despite the slight increases in tumor stage and age. As depicted in Figure 16, the cancer stage has a SHAP value of +0.16, which indicates a strong positive correlation with the recurrence of the DTC. Factors such as response, risk, and adenopathy favourably impact the recurrence, while tumor stage and age depict a slight negative influence. However, these two waterfall plots highlight that the variables, such as response to initial treatment, act as a key factor that drives the decision of thyroid cancer recurrence, despite the fact that variables such as tumor stage and adenopathy need careful consideration in some contexts.

However, despite the promising results and clear interpretations of this study, there are several notable limitations and challenges that should be considered. One of the major limitations of this study is the relatively small sample size, which contained 383 patients. This may restrict the model's capability in capturing hidden patterns and complex relationships, and impact the posterior estimation as well. In addition, the data were collected from a retrospective cohort in a single clinical centre, which limits the generalizability and fails to make interpretations on a diverse population. Furthermore, the absence of an external validation is also a major limitation of this study. Although cross-validation was conducted, the lack of external validation using a diverse dataset is a significant restriction of this study. Moreover, this study implemented several machine learning models for classification, which could be used as a benchmark in future studies. Compared to these machine learning models, the BNN model employed in this study, with varying prior distributions, performed well in classification. However, only six prior distributions were considered for the BNN model by the authors of this study, which may introduce a notable bias. In addition, the use of standard prior distributions may limit the exploration and integration of the expert knowledge. However, these limitations could be addressed strategically in future studies. For example, this study could be further expanded by incorporating a larger dataset from multicentres, thereby including a diverse population to increase the interpretability and generalizability of the model. In addition, the results obtained from this study can be confirmed, or any variations could be pointed out by considering an external validation dataset. Furthermore, this study was limited to a basic BNN architecture. Therefore, future studies could evaluate the performance of the BNN model with a deeper architecture by including additional hidden layers to capture more complex patterns. In addition, other Bayesian models, such as Bayesian logistic regression, could be used to compare the performance of the BNN model in thyroid cancer recurrence classification. Furthermore, by incorporating domain-informed priors, hierarchical priors, and expert opinions, this study can be further extended in order to provide robust results, thereby effectively classifying thyroid cancer recurrence.

## V. CONCLUSION

This study implemented both classical ML models and a BNN model with varying prior distributions to classify differentiated thyroid cancer recurrence using a dataset containing 383 patients with 16 variables. These models were trained using the complete set of variables and the reduced set obtained using the Boruta algorithm. SVM performed well before feature selection, obtaining an accuracy of 0.9481, while the LR model outperformed with an 0.9611 accuracy after feature selection. Notably, BNN with Normal (0, 10) prior distribution depicted stronger performance than both ML and BNN models before and after feature selection. It achieved an accuracy of 0.9870, offering robust uncertainty quantifications. Therefore, this model was selected as the best-performing model for DTC recurrence classification. Epistemic and aleatoric uncertainties were obtained for this model, reflecting confidence in the model's predictions. Furthermore, SHAP-based interpretation was performed to enhance the explainability of the BNN model by identifying the most influential features driving its predictions. Finally, it can be concluded that this study provides a comprehensive framework by combining feature selection, ML models, Bayesian models, uncertainty quantification and SHAP interpretations.